\documentclass{article}

\PassOptionsToPackage{square, numbers}{natbib}
\usepackage{amsthm}
\usepackage{diagbox}
\usepackage{algorithmicx} 
\usepackage{algorithm}
\usepackage{algpseudocode}
\usepackage{amstext}
\usepackage{graphicx, subfig}
\usepackage{array}
\usepackage[preprint]{neurips_2023}
\usepackage[utf8]{inputenc} % allow utf-8 input
\usepackage[T1]{fontenc}    % use 8-bit T1 fonts
\usepackage{url}            % simple URL typesetting
\usepackage{booktabs}       % professional-quality tables
\usepackage{amsfonts}       % blackboard math symbols
\usepackage{nicefrac}       % compact symbols for 1/2, etc.
\usepackage{microtype}      % microtypography
\usepackage{xcolor}         % colors
\usepackage{natbib}
\usepackage{MnSymbol}

\title{Local SGD Accelerates Convergence by Exploiting Second Order Information of the Loss Function}

\author{Linxuan Pan \\ Department of Electronic and Computer Engineering\\
The Hong Kong University of Science and Technology\\
Clear Water Bay, Kowloon, Hong Kong \\ \texttt{lpanak@connect.ust.hk} \And Shenghui Song \\ Department of Electronic and Computer Engineering\\
The Hong Kong University of Science and Technology\\
Clear Water Bay, Kowloon, Hong Kong \\  \texttt{eeshsong@ust.hk}  }
\begin{document}
	
	\maketitle
	\begin{abstract}
        With multiple iterations of updates, local statistical gradient descent (L-SGD) has been proven to be very effective in distributed machine learning schemes such as federated learning. In fact, many innovative works have shown that L-SGD with independent and identically distributed (IID) data can even outperform SGD. As a result, extensive efforts have been made to unveil the power of L-SGD. However, existing analysis failed to explain why the multiple local updates with small mini-batches of data (L-SGD) can not be replaced by the update with one big batch of data and a larger learning rate (SGD). In this paper, we offer a new perspective to understand the strength of L-SGD. We theoretically prove that, with IID data, L-SGD can effectively explore the second order information of the loss function. In particular, compared with SGD, the updates of L-SGD have much larger projection on the eigenvectors of the Hessian matrix with small eigenvalues, which leads to faster convergence. Under certain conditions, L-SGD can even approach the Newton method. Experiment results over two popular datasets validate the theoretical results.        
	\end{abstract}

\section{Introduction}	

\subsection{Background}
Due to the increasing concern on privacy and the  communication constraints, distributed machine learning scheme such as federated learning (FL) has attracted much attention recently from both academia and industry. The objective of FL is to train a global model, with weights $x$, by solving the following optimization problem
\begin{eqnarray} 
\text{min}_{x}F(x)=\frac{1}{m}\Sigma_{i=1}^{m} f_{i}(x) 
\end{eqnarray}
where $m$ denotes the number of clients and $f_{i}(\cdot)$ is the loss function of the $i$-th client. To solve the above problem without data sharing, innovative FL algorithms such as FedAvg \cite{mcmahan2017communication} have been proposed. With FedAvg, the training at clients are performed by local statistical gradient descent (L-SGD), where multiple iterations of local training are performed over mini-batches of data. The trained models from different clients are then uploaded and aggregated at the server by model weight averaging. With its privacy-preserving capability, FL has been applied to many practical applications  \cite{shyu2021systematic} \cite{sheller2020federated}. 

With the multiple iterations of training at the clients, L-SGD can speed up convergence and plays an extremely important role in reducing the communication cost of FL \cite{lin2018don} \cite{mcmahan2017communication} \cite{liu2020fedvision} \cite{hard2018federated}. As a result, L-SGD has attracted a lot of research attention. For example, \cite{stich2018local}  \cite{yu2019parallel} analysed the convergence rate of L-SGD for both convex and non-convex object functions with independent and identically distributed (iid) data. The corresponding analysis with Non-IID was performed in \cite{karimireddy2020scaffold} \cite{wang2021cooperative}.

Although there have been many engaging results on L-SGD, the fundamental reason for L-SGD to be able to accelerate convergence is still not well understood. In fact, the idea of L-SGD is not new and similar methods have also been utilized in centralized learning to accelerate the training of multiple local workers \cite{lin2018don} \cite{zhang2016parallel}. But, by far, we are still not clear why L-SGD can outperform SGD under certain circumstances. In this paper, we will try to answer this question from an optimization perspective. 

 \begin{algorithm}
		\caption{LOCAL SGD}
            \label{algorithm1}
		\begin{algorithmic}[1]
            %\Statex \textbf{input and symbol:} K number of local update,$P_{i}$underlying distribution for a dataset,$t$ update round,initialization $x_{0}$,learning rate $\eta$,$\xi^{t}_{i}$ a sampled mini-batch data of client i in round t,global loss function $F(x)$,$f_{i}(x)$ loss function of client i
            %\Statex \textbf{LOCAL SGD in iid setting}
            \For{t=0,$\cdots$,T-1}
		\For{i=1$\cdots$ m}
            \For{k=0,$\cdots$,K-1}
		\State $x_{t,i,k+1}=x_{t,i,k}-\eta \nabla f_{i}(x_{t,i,k},\xi_{t, i,k})$
            \EndFor
            \State Send $x_{t,i,k}$ to sever
            \EndFor
            \State $x_{t+1}=\frac{1}{m}\Sigma_{i=1}^{m}x_{t,i,k} $
            \EndFor
            \end{algorithmic}  		
\end{algorithm}

\section{Related Work}
To better review the related works, we first introduce the FL setting concerned in this paper. Consider a FL system where $m$ clients participate in total $T$ rounds of local training. In each round, all clients will perform $K$ mini-batches of local training where $\xi_{t,i,k}$ denotes the data utilized in the $k$-th iteration of local training in the $t$-th communication round by the $i$-th client. The corresponding L-SGD algorithm of the concerned system is shown in Algorithm \ref{algorithm1}.

\subsection{The myth of learning rate and number of local iterations}
There have been many works focusing on the analysis for the convergence rate of L-SGD, with both IID and Non-IID data. In Table \ref{table1}, we list the convergence rates derived by several important works. We also include the assumptions taken by different works, where LRK denotes the learning rate condition. 

\begin{table}[h]
\begin{center}
\caption{Convergence Analysis}
\label{table1}
\begin{tabular}{|c|c|c|c|c|}
\hline
Work & L-Lipschitz Continuous Gradient & Convexity & LRK & Convergence rate \\ \hline
\cite{karimireddy2020scaffold} & Yes & NC  & $K\eta \leq \frac{1}{16L}$  &  $\mathcal{O}(\frac{1}{T}+\frac{1}{\sqrt{mKT}})$ \\ \hline
\cite{yang2021achieving} &Yes & NC & $K\eta \leq \frac{1}{8L}$&$\mathcal{O}(\frac{1}{KT}+\frac{1}{\sqrt{mKT}})$ \\\hline
\cite{yu2019parallel} & Yes & NC & $\eta \leq \frac{1}{L}$ &  $\mathcal{O}(\frac{1}{\sqrt{mT}})$ \\\hline
\cite{khaled2020tighter} & Yes &  C & $\eta \leq \frac{1}{4L} $  & $\mathcal{O}(\frac{m^{\frac{3}{2}}}{\sqrt{KT}})$\\
\hline
\end{tabular}
\end{center}
\end{table}

The convergence bounds derived by \cite{karimireddy2020scaffold} \cite{yang2021achieving} \cite{khaled2019first} \cite{wang2021cooperative} \cite{yu2019parallel} \cite{khaled2020tighter}\cite{spiridonoff2020local} share a similar form and require a similar condition on the learning rate $\eta$ as follows
\begin{equation}
\label{condition}
    K*\eta \leq \frac{1}{NL}
\end{equation}
where $L$ is the Lipschitz constant and $N$ is a positive number greater than $1$.  This condition indicates that the learning rate $\eta$ and the number of local iterations $K$ are somehow equivalent, in the sense that the following two settings are supposed to provide the same  performance: 1) $\eta = C, K=1$ and 2) $\eta = \frac{C}{2}, K=2$. This, unfortunately, is far from being the truth, and has been falsified by many works. For example, the authors of \cite{lin2018don} 
\cite{mcmahan2017communication} showed that one update with a large learning rate can not replace L-SGD \cite{lin2018don}
\cite{mcmahan2017communication}. Furthermore, the authors of \cite{mcmahan2017communication} showed that, to obtain good performance, the number of local updates $K$ need to be set quite large, making the condition in (\ref{condition}) difficult to satisfy. 

\begin{figure}
\centering
   \includegraphics[width=.6\textwidth]{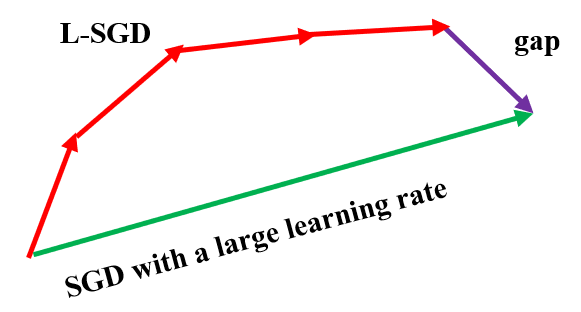}
\caption{Bound the gap}
\label{fig1}
  \end{figure}
  
The failure of existing theories in explaining 
the advantage of L-SGD over SGD comes from the adopted methodology. In particular, most existing works assumed that the direction of SGD is optimal and tried to bound the gap between the update by L-SGD and that by SGD, as shown in Fig. \ref{fig1}. In other words, extensive efforts have been devoted to bound the following term 
\begin{equation}
\label{bound}
    \Vert \text{update}_{SGD} - \text{update}_{LSGD}    \Vert^{2} = \Vert \nabla f(x_{t})K\eta -\frac{1}{m}\Sigma_{i=0}^{m-1}\Sigma_{k=0}^{K-1}  \nabla f_{i}(x_{t,i,k},\xi_{t,i,k})    \Vert^{2}.
\end{equation}
In particular, people hope that local SGD can generate update with a large projection on the (negative) direction of the gradient. For that purpose, people try to minimize the bounding error by restricting the size of the learning rate and the number of local updates. This effort, unfortunately, makes it even harder to unveil the effect of the local update.  

More importantly, the fundamental assumption that SGD gives the optimal descent direction is not necessarily true, because it only considers the first order information of the loss function. In fact, it is always more precise to use high order information to analyse the effect of the update on the loss function. In particular, the first and second order approximation for the update of the loss function can be given as follows
\begin{eqnarray}
&&\textbf{first order approximation:}  f(x_{t}-\eta \Delta ) \approx f(x_{t})-\eta \Delta^{T} \nabla f(x_{t}) \\
&& \textbf{second order approximation:}  f(x_{t}-\eta \Delta ) \approx f(x_{t})-\eta \Delta^{T} \nabla f(x_{t}) +\frac{1}{2}  \eta^{2} \Delta^{T} H(x_{t}) \Delta 
\label{approximation}
\end{eqnarray}
where $H(x_{t})$ denotes the Hessian matrix of the loss function at $x_t$. It is easy to argue that the second order approximation provides a more accurate characterization for the update of the loss function.  Unfortunately, it is normally difficult to directly bound the gradient by using higher order approximation. For example, when the commonly used L-Lipschitz continuous gradient assumption is applied to high order derivatives, it is difficult to capture the delicate geometry structure of complex models such as neural networks.

In this paper, we hope to investigate the higher order approximation from another perspective. In particular, we will leverage approximation theory to analyse the effect of local SGD. It will be theoretically shown that L-SGD can implicitly exploit the second order information of the loss function. Experiment results on popular datasets will validate the theoretical results. The approximation in this paper offers a new perspective in understanding the acceleration effect of L-SGD and provides insightful guidelines for the adjustment of the key hyper-parameters.  Furthermore, we will show by experiments that the practical values for the learning rate $\eta$ and the number of local update $K$ are far from those required by the popular learning rate assumption. 

\subsection{Our Contribution}
\begin{itemize}
    \item 1. In this paper, we investigate how local SGD can accelerate convergence. For that purpose, we theoretically prove that, with IID data, L-SGD can effectively explore the second order information of the loss function. In fact, under certain conditions, the convergence behavior of L-SGD can approach that of the Newton method. This is because the updates of L-SGD have much larger projection on the eigenvectors of the Hessian matrix with small eigenvalues, which leads to faster convergence than SGD. Experiment results over two popular datasets, i.e., MNIST and CIFAR-10, validate the theoretical results.
        
    \item 2. The results in this paper reveal the effects of the the learning rate $\eta$ and the number of local iteration $K$ on the update, which is different from the popular learning rate assumption. In fact, experiments demonstrated that, L-SGD can still accelerate convergence, even if the learning rate assumption widely adopted in the literature is not fulfilled.
\end{itemize}

\section{Background}

\subsection{Basic Assumptions and Notations}
With L-SGD, the $k$-th update process of the $i$-th client in the $t$-th round can be given by  
\begin{eqnarray}
    x_{t,i,k+1}=x_{t,i,k}-\eta \nabla f(x_{t,i,k},\xi_{t,i,k})
\end{eqnarray}
where $\nabla f(x_{t,i,k},\xi_{t,i,k})$ is a unbiased estimation for the gradient.
For ease of illustration, we further define the local update value as 
\begin{eqnarray}
\label{Dtik}
    \Delta_{t,i,k} = \eta \nabla f(x_{t,i,k},\xi_{t,i,k}). 
\end{eqnarray}
Then, the total local update for the $i$-th client in the $t$-th round can be given by
\begin{eqnarray}
\label{Dti}
    \Delta_{t,i} = \sum_{k=0}^{K-1}\Delta_{t,i,k},
\end{eqnarray}
and after aggregation, the global update in the $t$-th training round can be expressed as 
\begin{eqnarray}
    \Delta_{t} = \frac{1}{m}\sum_{i=1}^{m}\Delta_{t,i}.
    \label{avg}
\end{eqnarray}
  
Utilizing the Hessian matrix, we can approximate the gradient after $k$ local iterations as 
\begin{eqnarray}
    \nabla f(x_{t,i,k}) \approx \nabla f(x_{t,i,0})+ H(x_{t,i,0})(x_{t,i,k}-x_{t,i,0}),
\end{eqnarray}
where we can define the gradient estimation residue as follows. 

\textbf{Definition 1:} The gradient estimation residue is defined as 
\begin{eqnarray}
\label{ns}
    n^{s}(x_{t,i,k})=\nabla f(x_{t,i,k})-\nabla f(x_{t,i,0})-H(x_{t,i,0})(x_{t,i,k}-x_{t,i,0}).
\end{eqnarray}
Note that $n^{s}(x_{t,i,k})$ represents the higher order component ignored in the approximation. 

Given Definition 1, we can obtain 
\begin{eqnarray}
    \nabla f(x_{t,i,k})=\nabla f(x_{t,i,0})+H(x_{t,i,0})(x_{t,i,k}-x_{t,i,k-1}+x_{t,i,k-1}-x_{t,i,0})+ n^{s}(x_{t,i,k})
\end{eqnarray}
which can be further expressed as  
\begin{eqnarray}
    \nabla f(x_{t,i,k})=\nabla f(x_{t,i,k-1})+H(x_{t,i,0})(x_{t,i,k}-x_{t,i,k-1})+ n^{s}(x_{t,i,k})- n^{s}(x_{t,i,k-1})\label{fi}.
\end{eqnarray}
Taking expectation on both sides of (\ref{fi}), we can obtain 
\begin{eqnarray}
E\left[\nabla f(x_{t,i,k}) \right] = E\left[ \left( I_{d}-\eta H(x_{t,i,0})\right) \right] \nabla f(x_{t,i,k-1}))+E\left[ n^{s}(x_{t,i,k})- n^{s}(x_{t,i,k-1}) \right] \label{ni}
\end{eqnarray}
where $I_{d}$ denotes the unit matrix with $d$ dimensions.

\subsection{Local SGD can approach Newton method} 

\textbf{Assumption 1 (Bounded gradients):} For any model weight $x \in {R}^{d}$ and any sample data $\xi$ with a fixed batch size, the gradient norm is bounded
\begin{equation}
\Vert \nabla f(x,\xi) \Vert \leq G.
\end{equation}
Assumption 1 is commonly used and will also be validated by experiments later. With this assumption, we can bound the total local update for the i-th client in the t-th round as $\Vert \Delta_{t,i}\ \Vert \leq KG$. 
Given the norm of $\Delta_{t,i}$ is bounded, we can prove that $\Delta_{t,i}$ has a bounded variance, as shown in the following assumption. 

\textbf{Assumption 2 (Bounded variance):} The variance of $K$ local updates is bounded with
\begin{eqnarray}
    \text{Var}(\Delta_{t,i}) \leq  \sigma_{1}^{2}.
\end{eqnarray}

Next, we show that a large number of participating clients, $m$, is essential in accelerating the training. Given (\ref{avg}), it follows from Assumption 2 and the law of large numbers that when $m$ is large enough, $\Delta_{t}$ will converge to $E(\Delta_{t,i})$ with IID data. 

Next, we will determine $E(\Delta_{t,i})$.  
By substituting (\ref{ni}) into (\ref{Dtik}) and (\ref{Dti}) iteratively for $K$ times, we can obtain the following lemma.  

\textbf{Lemma 1:} The expectation of the total update for the i-th client in the t-th round can be given by 
\begin{eqnarray}
\label{EDelta}
E(\Delta_{t,i})=\sum_{k=0}^{K-1}(I_{d}-\eta H(x_{t,i,0}))^{k}\eta \nabla f(x_{t,i,0})+\sum_{k=0}^{K-1}(I_{d}-\eta H(x_{t,i,0}))^{K-1-k}\eta E\left[n^{s}(x_{t,i,k})\right].
\end{eqnarray}
The proof is given by Appendix \ref{PROOF1} in the supplementary materials. 

Note that the last term in (\ref{EDelta}) is related to $E\left[n^{s}(x_{t,i,k})\right]$. Next, we will look into the behavior of the gradient estimation residue, $n^{s}(x_{t,i,k})$. It follows from (\ref{ns}), that the residue $n^{s}(x_{t,i,k})$ should be very small if $x_{t,i,k}$ is very close to $x_{t,i,0}$. Surprisingly, we found that the norm of $E(n^{s}(x_{t,i,k}))$ is very small compared with the norm of $\nabla f(x_{t,i,0})$, even for a very large $k$. This phenomenon will be validated by experiments in Section 4.2.  

In this paper, we will show that the small value of $\Vert E(n^{s}(x_{t,i,k})\Vert$, compared with $\Vert \nabla f(x_{t,i,0}) \Vert$, is essential for accelerating convergence by L-SGD. Thus, we take the following assumption in this paper. 

\textbf{Assumption 3 (Small $n^{s}(x_{t,i,k})$):} During the training, we have 
\begin{eqnarray}
\label{Ens}
    \Vert E\left[n^{s}(x_{t,i,k}\right] \Vert<< \Vert \nabla f(x_{t,i,0}) \Vert.
\end{eqnarray}

This assumption will be validated by experiment results later. But, by far, we do not have a theory to explain this phenomenon. In fact, as shown by many experiments \cite{dauphin2014identifying} \cite{choromanska2015loss}, the loss surface of neural networks is not trivial. We conjecture that this property is a characteristic of the loss surface for nerual networks. One possible explanation is that the Hessian matrix of the loss function is very stable or there are some unknown dynamics about neural networks. A deeper explanation about this is beyond the scope of this article. 

In the following, we will further assume that the loss function has Lipschitz Gradient, which is widely adopted.

\textbf{Assumption 4: Lipschitz Gradient.} $f(x)$ is second-order differentiable. For any $x,y\in R^{d}$,we have 
\begin{eqnarray} 
    \Vert \nabla f(x)- \nabla f(y)\Vert \leq L\Vert x-y \Vert. 
\end{eqnarray}

By substituting (\ref{Ens}) into (\ref{EDelta}), we can obtain the following proposition when Assumption 3 holds.

\textbf{Proposition 1:} For a $\mu$ strongly convex loss function, if $\eta< \frac{1}{L}$ and $K$ is very large, we can obtain 
\begin{eqnarray}
\label{etaK}
      E\left[\Delta_{t,i}\right] &\approx&  H(x_{t})^{-1}(I_{d}-(I_{d}-\eta H(x_{t}))^{K})\nabla f(x_{t}) 
\end{eqnarray}
which can be further approximated as 
\begin{eqnarray}
 E\left[\Delta_{t,i}\right] &\approx&  H(x_{t})^{-1}\nabla f(x_{t}). 
\label{newton}
\end{eqnarray}
The proof is given in Appendix \ref{PROOF2} of the supplementary materials. 

\textbf{Remark 1:} The update shown in (\ref{newton}) is the same as that of Newton  method. This indicates that local SGD can implicitly utilize the second order information of the loss function to update the model.

\textbf{Remark 2:}  
It can be observed from (\ref{etaK}) that $K$ and  $\eta$ play totally different roles in the training process. In particular, $\eta$ should be limited by $1/L$ but not $K$. This is different from the common understanding in the literature that $\eta K$, i.e., the product of these two parameters, is limited by $1/LN$ where $N$ denotes a constant with $N \geq 1$. 
In the next subsection, we will show how local SGD can efficiently exploit the  Hessian matrix.

\subsection{Local SGD can implicitly use Hessian information}

To understand the influence of one update, we may leverage the Taylor expansion to approximate the improvement in the loss function. In the following, we will utilize the second order estimation (SOE) to approximate the update of the loss function. In particular, we can obtain 
\begin{equation}
\label{SOE}
f(x_{t}+\Delta_t)\approx f(x_{t})+\underbrace{\Delta_t^{T} \nabla f(x_{t}) +\frac{1}{2}  \Delta_t^{T} H(x_{t}) \Delta_t}_{u_{SOE}}.
\end{equation} 
In fact, the classic Newton method was derived based on the fact that the update $ H(x_{t})^{-1}\nabla f(x_{t})$ can minimize $u_{SOE}$.

In this paper, we will investigate how local SGD influences the SOE term. By taking the expectation of $u_{SOE}$ in (\ref{SOE}), we can obtain
\begin{eqnarray}
E\left[u_{SOE}\right] = E\left[\Delta_{t}\right] \nabla f(x_{t})+\frac{1}{2} E\left[ \Delta_{t}^{T}H(x_{t})\Delta_{t} \right]
\end{eqnarray}
which can be further expressed as  
\begin{eqnarray}
E\left[u_{SOE}\right] = E\left[\Delta_{t}\right]\nabla f(x_{t})+\frac{1}{2}E\left[\Delta_{t}\right]H(x_{t})E\left[\Delta_{t}\right]+
\frac{1}{2}E\left[(\Delta-E\left[\Delta_{t}\right])^{T}\right]H(x_{t})(\Delta-E\left[\Delta_{t}\right])
\label{ESOE}
\end{eqnarray}
As is implicitly adopted in the literature, we also assume the local updates $\Delta_{t,i}$ from different clients are independent, with the following assumption. \\
\textbf{Assumption 5: IID local update.} The local update of different clients in the t-th round $\Delta_{t,i}, i=1,...,m$ are statistically independent with each other.

By substituting (\ref{avg}) into the last term of (\ref{ESOE}), we can obtain.
\begin{eqnarray}
\frac{1}{2}E\left[(\Delta_{t}-E\left[\Delta_{t}\right])^{T} H(x_{t})(\Delta_{t}-E\left[\Delta_{t}\right]) \right]&\leq&\frac{L}{2}\sum_{i=1}^{m} \frac{Var(\Delta_{t,i})}{m^{2}}\\
&\leq& \frac{L\sigma_{1}^{2}}{m} \label{mless}
\end{eqnarray}
where the second line comes from Assumption 2. With a large $m$, we can then approximate (\ref{ESOE}) as 
\begin{eqnarray}
\label{ESOE1}
E\left[u_{SOE}\right] &\approx& E\left[\Delta_{t}\right]\nabla f(x_{t})+\frac{1}{2}E\left[\Delta_{t}\right]H(x_{t})E\left[\Delta_{t}\right].
\end{eqnarray}

Next, we will determine $E\left[\Delta_{t}\right]$. For that purpose, we will project $\Delta_{t}$ onto the direction of the eigenvectors of the Hessian matrix. Let $v_{l}, l=1,...d$ denote the eigenvectors of $H_{t}$ with corresponding eigenvalue $\lambda_{l}, l=1,...,d$. Further denote $w_l(y)$ as the projection of any vector $y$ on the eigenvector $v_{l}$.  Thus, the energy of $y$ along the direction of $v_{l}$ can be given by
\begin{equation}
e_{l}(y) = w_{l}(y)^{2}. 
\end{equation}

Based on Assumption 3 and the proof of Proposition 1, we can calculate the projection of $\Delta_{t,i}$ on the direction of $v_{l}$ and obtain 
\begin{eqnarray}
\label{wl}
E\left[w_{l}(\Delta_{t,i})\right] \approx  \frac{(1-(1-\eta \lambda_{l})^{K})w_{l}(\nabla f(x_{t}))}{\lambda_{l} }
\end{eqnarray}
where $\lambda_l \neq 0$. 

\textbf{Remark 3:} With a large $K$, (\ref{wl}) can be rewritten as  
\begin{eqnarray}
E(w_{l}(\Delta_{t,i})) \approx  \frac{w_{l}(\nabla f(x_{t}))}{\lambda_{l} },
\end{eqnarray}
which indicates that the energy of the local update on any eigenvector direction is equal to the energy of the gradient on that direction divided by the corresponding eigenvalue. As a result, the energy of the local update will concentrate on the directions with small eigenvalues. This result will be validated by experiments in Section 4.

By combining all the projections on $d$ dimensions, we can obtain $E\left[\Delta_{t,i}\right]$. Given (\ref{avg}), we know that $E\left[\Delta_{t}\right] = E\left[\Delta_{t,i}\right]$. Finally, by substituting $E\left[\Delta_{t}\right]$ into (\ref{ESOE1}), we can obtain  
\begin{eqnarray}
E(u_{SOE})\approx \sum_{l=1}^{d} s_{l}
\end{eqnarray}
where 
\begin{eqnarray}
\label{sl}
s_{l}=\left \{ \begin{aligned}
-K\eta*e_{l}(\nabla f(x_{t})) \eta \quad &\lambda_{l}=0 \\ 
\frac{-(1-(1-\lambda_{l}\eta)^{2K})e_{l}(\nabla f(x_{t}))}{2\lambda_{l}}\quad & \lambda_{l}\neq 0 .
\end{aligned} \right.
\end{eqnarray}
$s_{l}$ can be regarded as the contribution of the update to $u_{SOE}$ on the $v_{l}$ direction.  

\textbf{Remark 4:} It can be observed from (\ref{sl}) that the impact of the learning rate $\eta$ and the number of local update $K$ is not equivalent. Furthermore, when $\lambda_l >0$, if $\eta < 1/L < 1/\lambda_l$, increasing $K$ will decrease $s_l$, which in turn reduces $u_{SOE}$ and the loss function. However, given $K$ is at the exponent, its effect may saturate very soon. This reveals the impact of $\eta$ and $K$, which is different from the current understanding based on the popular learning rate assumption. 

\section{Experiment}

\subsection{Experiment Settings}
To show why local SGD can accelerate convergence, we performed experiments on two popular datasets, i.e., 
MNIST and CIFAR-10, with different neural network models.
In the experiments, we assume there are in total 100 clients and the data of different clients are independent and identically distributed.  

Note that our purpose in this paper is not to achieve the best performance but to unveil the reason why local SGD can accelerate convergence. Thus, we adopted two simple neural networks for the two datasets, respectively. For MNIST, we used a simple fully connected neural network model with only one hidden layer, and trained it for 300 round on MNIST. For CIFAR-10, we used a 4-layer CNN with 2 convolutional layers and two fully connected layers, and trained it for 120 round. We fixed the learning rate to be $\eta=0.01$ and batch size to be 10 for both datasets. The number of local iterations for the two datasets was set to be $K=300$ and $K=200$, respectively. 

For comparison purposes, we also trained these two models by SGD with a batch size of 1000 and learning rate $\eta \in \{0.05,0.1,0.2,0.3\}$. 

\begin{figure}
		\centering	\includegraphics[width=.71\textwidth]{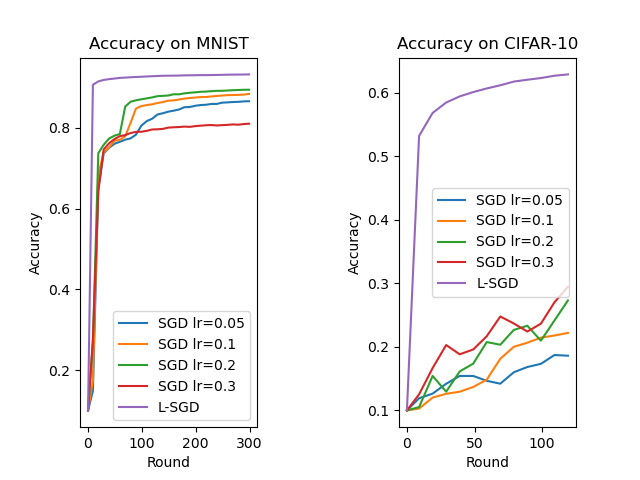}
		\caption{Performance comparison between L-SGD and SGD}
            \label{fig2}
	\end{figure}

%\textbf{1:} LRK condition doesn't hold in many circumstances. The gap is very big. 

%\textbf{2:} As we predict,that the energy of the SOE concentrates on the eigenvector direction  of Hessian matrix whose  eigenvalue is very is near zero.

%\textbf{3:} Assumption 3 hold $\Vert E(n^{s}(x_{t,i,K-1})) \Vert << \Vert \nabla f(x_{t,i,K-1}) \Vert $.Also,through our experiment we can view one interesting 
%phenomenon,that the variance of $n^{s}(x_{t,i,K-1})$ is nor small.We argue that this phenomenon implicitly shows the dynamic and loss surface of neural network may have 
%special property.

\subsection{Learning rate assumption doesn't hold}
We first compare the performance of L-SGD with that of SGD on the two datasets and the results are shown in Fig. \ref{fig2}. It can be observed that L-SGD can accelerate convergence compared with SGD. This phenomenon has also be reported by other works.

As mentioned above, existing works proved the convergence of L-SGD with an assumption on the learning rate $K\eta \leq \frac{1}{L N}$, where $L$ is larger than the norm of the largest eigenvalue of the Hessian matrix.  Now, we show that L-SGD can achieve convergence acceleration without satisfying the learning rate assumption. With $\eta = 0.01$ and $K=200,300$ in our experiments, we can obtain $K \eta = 2, 3$.  In Table \ref{table2}, we show the range of the eigenvalues with different experiment settings at communication round 10, 60, and 90. It can be easily checked that the learning rate assumption does not hold, but the training performance shown in Fig. \ref{fig2} demonstrates the effectiveness of L-SGD in accelerating convergence.   

\begin{table}[h]
\begin{center}\caption{Range of Eigenvalues at Different Round}
\label{table2}
\begin{tabular}{|c|c|c|}
\hline
Dataset & Round &   Eigenvalues Range \\
\hline
MNIST   &    10  &  [-4.869e-10, 1.558]\\
\hline
MNIST   &    60  &    [-1.266e-10, 1.155]\\
\hline
MNIST  &     100  &    [-1.755-10, 1.068]\\
\hline
CIFAR-10  &   10   &   [-0.338, 34.113] \\
\hline
CIFAR-10  &  60   &    [-0.163, 31.492]\\
\hline
CIFAR-10  &  100  &    [-0.126, 31.586]  \\
\hline
\end{tabular}
\end{center}
\end{table}

\begin{figure}
		\centering
	\includegraphics[width=.6\textwidth]{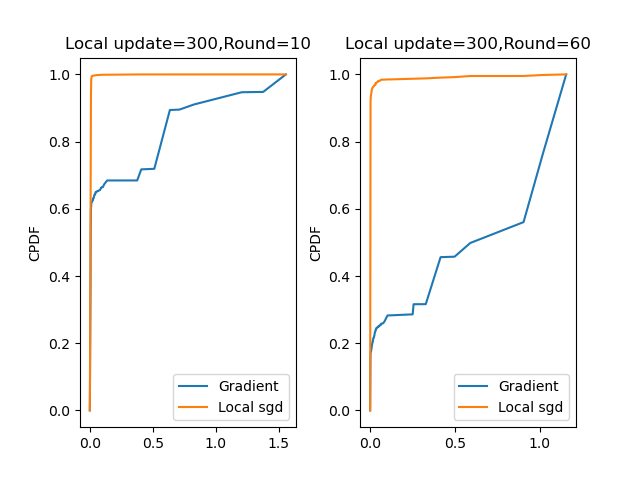}
		\caption{CPDF at different communication round on MNIST dataset}
         \label{MC}
	\end{figure}
\begin{figure}
		\centering	\includegraphics[width=.6\textwidth]{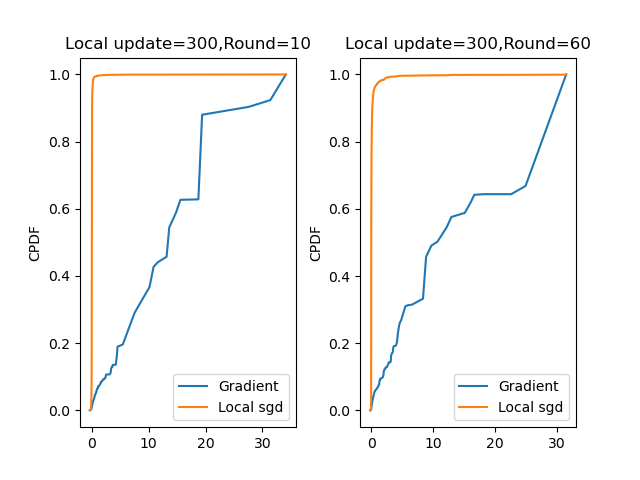}
		\caption{CPDF at different communication round on CIFAR-10 dataset}
         \label{CC}
	\end{figure}
 
\subsection{Energy distribution of the local update} 
As discussed in Remark 3, the fundamental reason that L-SGD can accelerate convergence may come from the fact that the energy of the local update concentrates on the eigen-directions whose eigen-values are very small. To illustrate the energy distribution, we define the Cumulative Power Distribution Function (CPDF) as follows  
\begin{eqnarray}
F_{S_{l}=\{l:(x>=\lambda_{l})\}}(x)=\frac{\sum_{l\in S_{l}} w_{l}(x)^{2}}{\sum_{l=1}^{d} w_{l}(x)^{2}} 
\end{eqnarray}
where $w_{l}$ has been defined in Section 3.3. In particular, 
CPDF illustrates the distribution of the energy on different eigen directions. In Fig. \ref{MC}, we compare the CPDF for L-SGD and GD on the MNIST dataset at round 10 and 60. The corresponding comparison on the CIFAR-10 dataset is reported in Fig. \ref{CC}. It can be observed that, for both experiments, the energy of L-SGD indeed concentrates on the directions with small eigenvalues, which is not true for the gradient. This observation agrees with Remark 3. More experiment results will be shown in the supplementary materials.

\subsection{Validation of Assumption 3}
In this section, we will validate the correctness of Assumption 3. For that purpose, we estimate the value of $E\left[n^{s}(x_{t,i,k})\right]$ and report the ratio $\frac{\Vert \nabla f(x_{t,i,0}) \Vert}{\Vert \hat{E\left[n^{s}(x_{t,i,k})\right]} \Vert}$, where $\hat{E\left[n^{s}(x_{t,i,k})\right]}$ denotes the estimation of $E\left[n^{s}(x_{t,i,k})\right]$.  The estimation of $E\left[n^{s}(x_{t,i,k})\right]$ is based on (\ref{ns}) and the results are reported in Table \ref{table3}. In particular, we picked several combinations between $t={10, 100}$ and $k={10, 30, 100, 300}$, and picked $x_{t,i,0}$ as the initial point. For each combination, we ran 500 trials and took the average to calculate $\hat{E\left[n^{s}(x_{t,i,k})\right]}$.    
It can be observed that, for all experiment, $E\left[n^{s}(x_{t,i,k})\right]$ is very small compared with $\Vert \nabla f(x_{t,i,0}) \Vert$.

\begin{table}[h]
\begin{center}\caption{Validation of Assumption 3 at different round and iteration}
\label{table3}
\begin{tabular}{|c|c|c|}
\hline
Round t &  Iteration k  &  $\frac{\Vert \nabla f(x_{t,i,0})\Vert}{\Vert   \hat{E(n^{s}(x_{t,i,k}))} \Vert}$ \\
\hline
10  &    10 &  167.3\\
\hline
10   &    30  &    85.1\\
\hline
10  &     100  &    19.5\\
\hline
10  &   300  &   10.2 \\
\hline
100  &  10   &    48.2\\
\hline
100  &  30 &    50.0 \\
\hline
100  &  100 &    14.4\\
\hline
100  &  300 &    7.36\\
\hline
\end{tabular}
\end{center}
\end{table}

\section{Conclusion}
In this paper, we investigated the reason why local SGD can speed up convergence for distributed learning schemes, such as federated learning. By taking the second order information into consideration, we first showed that, under certain conditions, local SGD can approach the Newton method. By investigating the energy projection of the local update on different eigen-directions of the Hessian matrix, we illustrate how the second order information of the loss function is utilized by local SGD to accelerate convergence. To be more specific, we showed that the fundamental reason for local SGD to outperform SGD/GD comes from the fact that the update by local SGD concentrates its energy on the eigen-directions of the Hessian matrix with small eigenvalues. The approximation result in this paper offers a new perspective to understand the power of local SGD and extensive future works are needed to fully understand the novel behavior of neural networks.    
\bibliographystyle{plain}
\bibliography{ref.bib}

\newpage
\appendix
\section*{APPENDIX}

In this appendix, we will first provide the proofs for Lemma 1 and Proposition 1. Then, we will provide additional experiment results on MNIST and CIFAR-10 datasets to validate Remark 3. Finally, we will discuss the key innovation and limitation of this work, and provide details of the machine learning models utilized in the experiments. 

\setcounter{section}{0}
\section{Proof}

\subsection{Proof of Lemma 1.} \textbf{Lemma 1:} The expectation of the total update for the i-th client in the t-th round can be given by 
\begin{eqnarray}
E\left[\Delta_{t,i}\right]=\sum_{k=0}^{K-1}(I_{d}-\eta H(x_{t,i,0}))^{k}\eta \nabla f(x_{t,i,0})+\sum_{k=0}^{K-1}(I_{d}-\eta H(x_{t,i,0}))^{K-1-k}\eta E\left[n^{s}(x_{t,i,k})\right]
\label{appL1}
\end{eqnarray}
%\textit{Proof.}
\begin{proof}
\label{PROOF1}
According to the definition of $n^{s}(x_{t,i,k})$ in (11), we have 
\begin{eqnarray}
\nabla f(x_{t,i,k})&=&\nabla f(x_{t,i,0}) +H(x_{t,i,0})(x_{t,i,k}-x_{t,i,0})+n^{s}(x_{t,i,k}).
\label{app35}
\end{eqnarray}
By adding and subtracting $H(x_{t,i,0})(x_{t,i,k-1})$ and $n^{s}(x_{t,i,k-1})$ on the right hand side (RHS) of (\ref{app35}), we can obtain
\begin{eqnarray}
\nabla f(x_{t,i,k}) &=&\nabla f(x_{t,i,0}) + H(x_{t,i,0})(x_{t,i,k}-x_{t,i,k-1})\nonumber\\&+&H(x_{t,i,0})(x_{t,i,k-1}-x_{t,i,0})+n^{s}(x_{t,i,k})+n^{s}(x_{t,i,k-1})-n^{s}(x_{t,i,k-1}). 
\label{app36}
\end{eqnarray}
By applying (\ref{app35}) on the RHS of (\ref{app36}), we can further obtain
\begin{eqnarray}
\nabla f(x_{t,i,k})=  \nabla f(x_{t,i,k-1})+  H(x_{t,i,0})(x_{t,i,k}-x_{t,i,k-1})+n^{s}(x_{t,i,k})-n^{s}(x_{t,i,k-1})
\label{app37}
\end{eqnarray}
for $k>=1$. By taking the expectation on both sides of (\ref{app37}), we have
\begin{eqnarray}
\label{expect}
E\left[\nabla f(x_{t,i,k})\right]=(I_{d}-\eta H(x_{t,i,0}))E\left[\nabla f(x_{t,i,k-1})\right] +E\left[n^{s}(x_{t,i,k})\right]-E\left[n^{s}(x_{t,i,k-1})\right],
\end{eqnarray}
which gives the iterative relation between $E\left[\nabla f(x_{t,i,k})\right]$ and $E\left[\nabla f(x_{t,i,k-1})\right]$. 

By applying this iterative relation on the RHS of (\ref{expect}) for $k-1$ times, we can obtain
\begin{eqnarray}
\label{core}
E\left[\nabla f(x_{t,i,k})\right]&=&(I_{d}-\eta H(x_{t,i,0}))^{k}E\left[\nabla f(x_{t,i,0})\right]\nonumber \\
&+&\sum_{j=1}^{k} (I_{d}-\eta H(x_{t,i,0}))^{k-j}(E\left[n^{s}(x_{t,i,j})\right]-E\left[n^{s}(x_{t,i,j-1})\right]). 
\end{eqnarray}

For the cases with $K=1$ and $K=2$, Lemma 1 can be proved by substituting the definition of $n^{s}(x_{t,i,k})$ into (\ref{appL1}).  For $K\geq 3$, we have  
\begin{eqnarray}
\label{last2}
E\left[\Delta_{t,i}\right]=\sum_{k=0}^{K-1} \eta E\left[ \nabla f(x_{t,i,k})\right]. 
\end{eqnarray}
By substituting (\ref{core}) into (\ref{last2}), we can further obtain
\begin{eqnarray}
E\left[\Delta_{t,i}\right] &=& \sum_{k=0}^{K-1}(I_{d}-\eta H(x_{t,i,0}))^{k}\eta \nabla f(x_{t,i,0}) \nonumber \\
&+& \sum_{k=1}^{K-1}\sum_{j=1}^{k}  \eta (I_{d}-\eta H(x_{t,i,0}))^{k-j}(E\left[n^{s}(x_{t,i,j})\right]-E\left[n^{s}(x_{t,i,j-1})\right])
\label{app41}
\end{eqnarray}
where the second term on the RHS of (\ref{app41}) can be separated to two double summations to obtain
\begin{eqnarray}
E\left[\Delta_{t,i}\right]
&=&  \sum_{k=0}^{K-1}(I_{d}-\eta H(x_{t,i,0}))^{k}\eta \nabla f(x_{t,i,0})+\sum_{j=1}^{K-1} \sum_{k=j}
^{K-1} (I_{d}-\eta H(x_{t,i,0}))^{k-j}\eta E\left[n^{s}(x_{t,i,j})\right] \nonumber \\ 
&-&\sum_{j=0}^{K-2} \sum_{k=j+1}^{K-1} (I_{d}-\eta H(x_{t,i,0}))^{k-1-j} \eta  E\left[n^{s}(x_{t,i,j})\right].  
\label{app42}
\end{eqnarray}
By separating the second and third terms in (\ref{app42}) to two terms, we can further obtain 
\begin{eqnarray}
E\left[\Delta_{t,i}\right]
&=& \sum_{k=0}^{K-1}(I_{d}-\eta H(x_{t,i,0}))^{k}\eta \nabla f(x_{t,i,0})+\sum_{k=1}^{K-1} (I_{d}-\eta H(x_{t,i,0}))^{k-1} \eta  E\left[n^{s}(x_{t,i,0})\right] 
\nonumber \\
&+& \left(\sum_{j=1}^{K-2} \sum_{k=j}^{K-1} (I_{d}-\eta H(x_{t,i,0}))^{k-j}-\sum_{j=1}^{K-2} \sum_{k=j}^{K-2} (I_{d}-\eta H(x_{t,i,0}))^{k-j}\right) \eta E\left[n^{s}(x_{t,i,j})\right] \nonumber \\
&+& \eta E\left[n^{s}(x_{t,i,K-1})\right].
\end{eqnarray}
After some mathematical manipulations, we have
\begin{eqnarray}
E\left[\Delta_{t,i}\right] &=& \sum_{k=0}^{K-1}(I_{d}-\eta H(x_{t,i,0}))^{k}\eta \nabla f(x_{t,i,0})+\sum_{k=1}^{K-1} (I_{d}-\eta H(x_{t,i,0}))^{k-1} \eta  E\left[n^{s}(x_{t,i,0})\right] \nonumber \\
&+& \sum_{k=1}^{K-1}(I_{d}-\eta H(x_{t,i,0}))^{K-1-k}\eta E\left[n^{s}(x_{t,i,k})\right].
\label{app44}
\end{eqnarray}
By the definition of $n^{s}(x_{t,i,k})$, we know
\begin{eqnarray}
    n^{s}(x_{t,i,0})=\vec{0}.
\end{eqnarray}
It follows that the second term on the RHS of (\ref{app44}) is zero, and we can rewrite (\ref{app44}) in a more elegant form as 
\begin{eqnarray}
E\left[\Delta_{t,i}\right]&=&\sum_{k=0}^{K-1}(I_{d}-\eta H(x_{t,i,0}))^{k}\eta \nabla f(x_{t,i,0})+\sum_{k=0}^{K-1}(I_{d}-\eta H(x_{t,i,0}))^{K-1-k}\eta E\left[n^{s}(x_{t,i,k})\right] \nonumber
\end{eqnarray}
which completes the proof of Lemma 1. 
\end{proof}

\subsection{Proof of Proposition 1.} \textbf{Proposition 1:} For a $\mu$ strongly convex loss function, if $\eta< \frac{1}{L}$ and $K$ is very large, we can obtain 
\begin{eqnarray}
      E\left[\Delta_{t,i}\right] &\approx&  H(x_{t})^{-1}(I_{d}-(I_{d}-\eta H(x_{t}))^{K})\nabla f(x_{t}) 
      \label{app46}
\end{eqnarray}
which can be further approximated as 
\begin{eqnarray}
 E\left[\Delta_{t,i}\right] &\approx&  H(x_{t})^{-1}\nabla f(x_{t}). 
\end{eqnarray}
%\textit{Proof.}  
\begin{proof}
\label{PROOF2}
By Lemma 1, we have 
\begin{eqnarray}
E\left[\Delta_{t,i}\right]&=&\sum_{k=0}^{K-1}(I_{d}-\eta H(x_{t}))^{k}\eta \nabla f(x_{t})+\sum_{k=0}^{K-1}(I_{d}-\eta H(x_{t}))^{K-1-k}\eta E\left[n^{s}(x_{t,i,k})\right]\\
&=&\sum_{k=0}^{K-1} (I_{d}-\eta H(x_{t}))^{k}\eta(\nabla f(x_{t})+E\left[n^{s}(x_{t,i,K-1-k})\right]).
\end{eqnarray}
With Assumption 3, we can approximate $E\left[\Delta_{t,i}\right]$ as
\begin{eqnarray}
E\left[\Delta_{t,i}\right] \approx \sum_{k=0}^{K-1}(I_{d}-\eta H(x_{t}))^{k}\eta \nabla f(x_{t}). 
\end{eqnarray}
By the sum of geometric series, we can further obtain 
\begin{eqnarray}
E\left[\Delta_{t,i}\right] %&\approx&\frac{I_{d}-(I_{d}-\eta %H(x_{t}))^{K}}{\eta H(x_{t})}\eta \nabla % f(x_{t})\\
\approx  H(x_{t})^{-1}(I_{d}-(I_{d}-\eta H(x_{t}))^{K})\nabla f(x_{t}). 
\end{eqnarray}
The Hessian matrix $H(x_{t})$ can be diagonilized as 
\begin{eqnarray}
H(x_{t})&=&Q^{T}\begin{bmatrix} 
    \lambda_{1} & 0 & \cdots & 0\\
   0 & \lambda_{2} & \cdots & 0 \\
   \vdots & \vdots &\ddots & \vdots\\
   0 & 0 & \cdots & \lambda_{d}
	\end{bmatrix}
 Q
\end{eqnarray}
where $Q$ is a orthogonal-normal matrix and $\lambda_{i}, i=1, ..., d$ represents the eigenvalue of $H(x_{t})$. Thus, we can compute $I_{d}-\eta H(x_{t})$ as 
\begin{eqnarray}
I_{d}-\eta H(x_{t})&=&Q^{T}I_{d}Q-Q^{T}\begin{bmatrix} 
    \eta\lambda_{1} & 0 & \cdots & 0\\
   0 & \eta\lambda_{2} & \cdots & 0 \\
   \vdots & \vdots &\ddots & \vdots\\
   0 & 0 & \cdots & \eta \lambda_{d}
   \end{bmatrix} Q \\
   &=&Q^{T}\begin{bmatrix} 
    1-\eta\lambda_{1} & 0 & \cdots & 0\\
   0 & 1-\eta\lambda_{2} & \cdots & 0 \\
   \vdots & \vdots &\ddots & \vdots\\
   0 & 0 & \cdots & 1-\eta\lambda_{d}
   \end{bmatrix} Q. 
\end{eqnarray}

\begin{figure}
		\centering	\includegraphics[width=.65\textwidth]{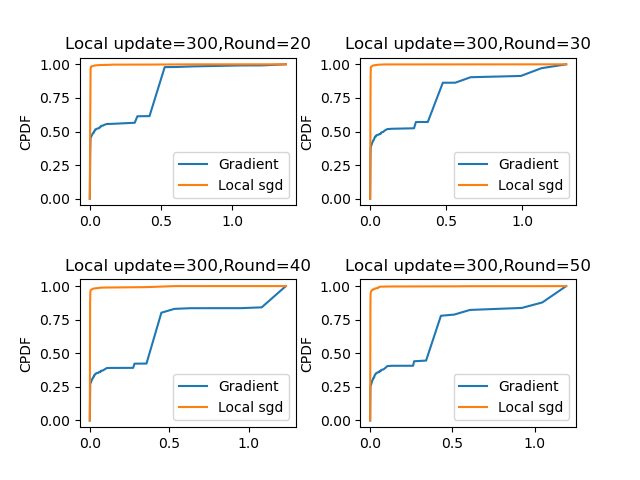}
		\caption{CPDF comparison between L-SGD and SGD on MNIST (Round 20-50)}
            \label{mnist20}
\end{figure}

Similarly, we can compute $I_{d}-(I_{d}-\eta H(x_{t}))^{K}$ as 
\begin{eqnarray}
I_{d}-(I_{d}-\eta H(x_{t}))^{K}&=&Q^{T}\begin{bmatrix} 
    1-(1-\eta\lambda_{1})^{K} & 0 & \cdots & 0\\
   0 & 1-(1-\eta\lambda_{2})^{K} & \cdots & 0 \\
   \vdots & \vdots &\ddots & \vdots\\
   0 & 0 & \cdots & 1-(1-\eta\lambda_{d})^{K}
   \end{bmatrix} Q. \nonumber
\end{eqnarray}
As $\eta < \frac{1}{L}$ and $\lambda_{i}\geq \mu$, we can obtain $\eta \mu \leq \eta \lambda_{i} <1 $ for $i=1, \cdots, d$. 
As a result, we know that as $K \rightarrow +\infty$, we have 
\begin{eqnarray}
1-(1-\eta\lambda_{1})^{K} \rightarrow 1.
\end{eqnarray}
Similarly, when $K\rightarrow +\infty$, we can obtain
\begin{eqnarray}
I_{d}-(I_{d}-\eta H(x_{t}))^{K} \approx I_{d}.
\label{app55}
\end{eqnarray}
By substituting (\ref{app55}) into (\ref{app46}), we complete the proof. \end{proof}

\begin{figure}
		\centering	\includegraphics[width=.65\textwidth]{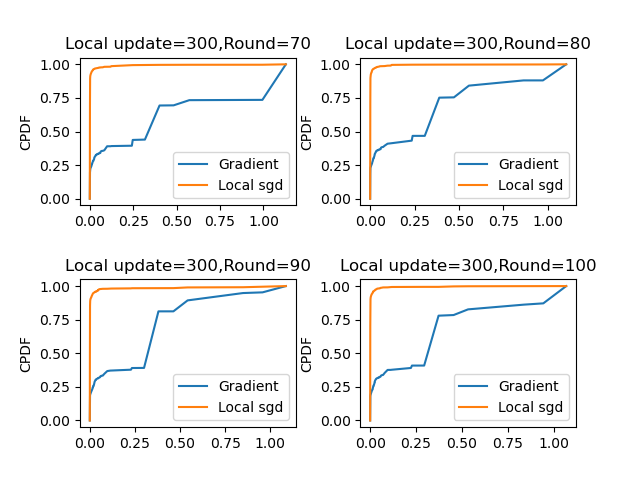}
		\caption{CPDF comparison between L-SGD and SGD on MNIST (Round 70-100)}
            \label{mnist70}
\end{figure}

\section{Additional Experiments to Validate Remark 3}
\begin{figure}
		\centering	\includegraphics[width=.65\textwidth]{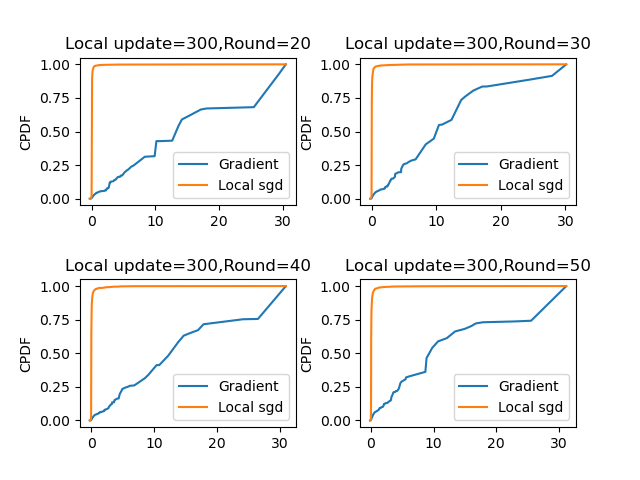}
		\caption{CPDF comparison between L-SGD and SGD on CIFAR-10 (Round 20-50)}
            \label{cifar20}
\end{figure}

\begin{figure}
		\centering	\includegraphics[width=.65\textwidth]{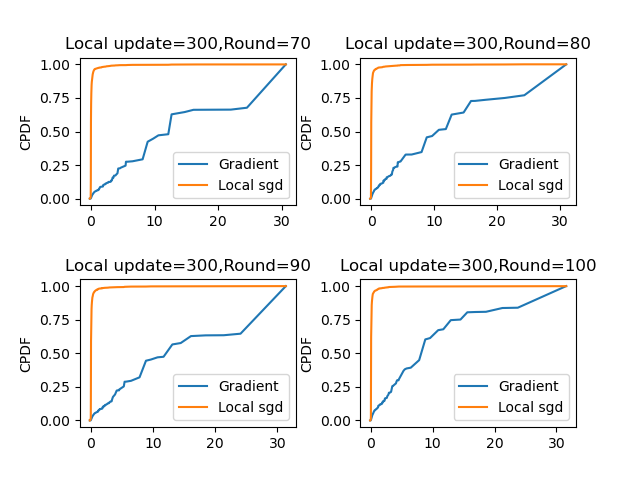}
		\caption{CPDF comparison between L-SGD and SGD on CIFAR-10 (Round 70-100)}
            \label{cifar70}
\end{figure}

In the main paper, we compared the CPDF for L-SGD and GD at rounds 10 and 60, on the MNIST and CIFAR-10 datasets. In this section, we further show the results at rounds 20, 30, 40, 50, 70, 80, 90, 100 on both datasets. The results for MNIST are shown in Fig. \ref{mnist20} and Fig. \ref{mnist70}, and those for CIFAR-10 are shown in Fig. \ref{cifar20} and Fig. \ref{cifar70}. 

It can be observed from all the figures that the energy of the local update by L-SGD concentrates on the eigen-directions of the Hessian matrix with small eigenvalues, which is not true for GD. This provides further validations for Remark 3.  Furthermore, this observation offers an interesting insights regarding the update direction of L-SGD, which can not be exposed based on the conventional learning rate condition.

\section{Key Innovation and Limitation of This Work}

\textbf{Key Innovation.} The key contribution of this work is the development of an intuitive model to investigate the impact of the learning rate $\eta$ and the number of local iteration $K$ on L-SGD. This model can be used to explain the direction of the local update by L-SGD, which can not be unveiled by the existing analysis based on the learning rate condition. In particular, the theoretical analysis in this work, though through approximation, predicts one important phenomenon, i.e., the energy of the local update concentrates on the eigen-directions of the Hessian matrix with extremely small eigenvalues, which was validated by extensive experiment results. This phenomenon explains why L-SGD can effectively reduce the loss function and thus accelerate convergence. Although the result in this work is not  enough to fully explain why L-SGD can accelerate the training of neural networks, Remark 3 can be regarded as a small step towards more advanced theories to characterize the behavior of local updates, which hopefully will be developed in the near future. 

\textbf{Discussion about Assumption 3:} Assumption 3 is very unexpected and not commonly utilized. In this paper, we verified Assumption 3 by experiment results with simple machine learning models, but it may not hold for complex model architectures, especially when the number of local updates $K$ is very large. Obviously, we still have a long way to go before we can fully understand the behavior of L-SGD. However, the theoretical analysis based on Assumption 3 is very intuitive and such a view offers a promising direction to explore the dynamics of L-SGD. 

%We still think this is an important clues for understanding L-SGD in the future. Because when we mention GD and SGD we know update model following the direction of gradient. However, in the past, we don't understand what kind of direction L-SGD move to and don't have any insight in that aspect. Our work offers a clear insight for the direction of L-SGD update. The loss surface and dynamic of neural network are both nontrivial, so we may have a long way to go.
\section{Model Architecture}
Details of the machine learning models utilized in this work are shown in Table \ref{table4}. For fully connected layer (FC), we list the parameter sequence (input dimension, output dimension). For convolutional layer (Conv2d), we list the parameter sequence (input dimentsion, output dimension, kernel size, stride, padding). For max pooling layer (Maxpool), we list the parameter sequence (kernel, stride). RELU represents rectified linear unit activation function layer.
 
\begin{table}[h]
\begin{center}\caption{Model Details}
\label{table4}
\begin{tabular}{|c|c|c|}
\hline
Model &Layer &  Details    \\
\hline
MLP & 1  &  FC(784, 10), RELU\\
\hline
CNN & 1 & Conv2d(3, 5, 5, 1, 0), RELU, Maxpool(2,2)\\
\hline
CNN & 2 & Conv2d(5, 10, 5, 1, 0), RELU, Maxpool(2,2)\\
\hline
CNN & 3 & FC(250, 50), RELU\\
\hline
CNN & 4 & FC(50, 10), RELU \\
\hline
\end{tabular}
\end{center}
\end{table}

\end{document}